\documentclass{article}

    \PassOptionsToPackage{numbers, compress}{natbib}

\usepackage[preprint]{neurips_2025}

\usepackage[utf8]{inputenc} 
\usepackage[T1]{fontenc}    
\usepackage{hyperref}       
\usepackage{url}            
\usepackage{booktabs}       
\usepackage{amsfonts}       
\usepackage{nicefrac}       
\usepackage{microtype}      
\usepackage{xcolor}         
\usepackage{graphicx}
\usepackage{amsmath}

\usepackage[normalem]{ulem}
\usepackage{multirow}
\usepackage{colortbl}
\newcommand{\tabincell}[2]{\begin{tabular}{@{}#1@{}}#2\end{tabular}}
\usepackage{enumitem}

\title{Diff-MM: Exploring Pre-trained Text-to-Image Generation Model for Unified Multi-modal Object Tracking}

\author{%
  Shiyu~Xuan, Zechao~Li, Jinhui~Tang \\
  School of Computer Science and Engineering, Nanjing University of Science and Technology \\
  \texttt{shiyu\_xuan@njust.edu.cn,zechao.li@njust.edu.cn,jinhuitang@njust.edu.cn} \\
}

\begin{document}

\maketitle

\begin{abstract}
    Multi-modal object tracking integrates auxiliary modalities such as depth, thermal infrared, event flow, and language to provide additional information beyond RGB images, showing great potential in improving tracking stabilization in complex scenarios. Existing methods typically start from an RGB-based tracker and learn to understand auxiliary modalities only from training data. Constrained by the limited multi-modal training data, the performance of these methods is unsatisfactory. To alleviate this limitation, this work proposes a unified multi-modal tracker Diff-MM by exploiting the multi-modal understanding capability of the pre-trained text-to-image generation model. Diff-MM leverages the UNet of pre-trained Stable Diffusion as a tracking feature extractor through the proposed parallel feature extraction pipeline, which enables pairwise image inputs for object tracking. We further introduce a multi-modal sub-module tuning method that learns to gain complementary information between different modalities. By harnessing the extensive prior knowledge in the generation model, we achieve a unified tracker with uniform parameters for RGB-N/D/T/E tracking. Experimental results demonstrate the promising performance of our method compared with recently proposed trackers, e.g., its AUC outperforms OneTracker by 8.3\% on TNL2K.
\end{abstract}

\section{Introduction}
\label{sec:intro}

Given a sequence of video frames and the initial bounding box of an object, single object tracking aims to localize the object in each frame. As a fundamental task in computer vision, it has broad applications in fields ranging from autonomous driving~\cite{geiger2012we}, smart city~\cite{xing2010multiple} to embodied AI~\cite{wang2024embodiedscan}. Despite the impressive progress, RGB-based object tracking still struggles in some challenging scenarios, \emph{e.g.}, low illumination, fast-moving object. To address these limitations, multi-modal tracking~\cite{depthtrack,lasher,wang2021towards,visevent} integrates auxiliary modalities (\emph{e.g.}, \textbf{D}epth, \textbf{E}vent flow, \textbf{T}hermal infrared, \textbf{N}atural language) to complement RGB images, thereby enhancing tracking robustness.

\begin{figure}
  \centering
   \includegraphics[width=1.0\linewidth]{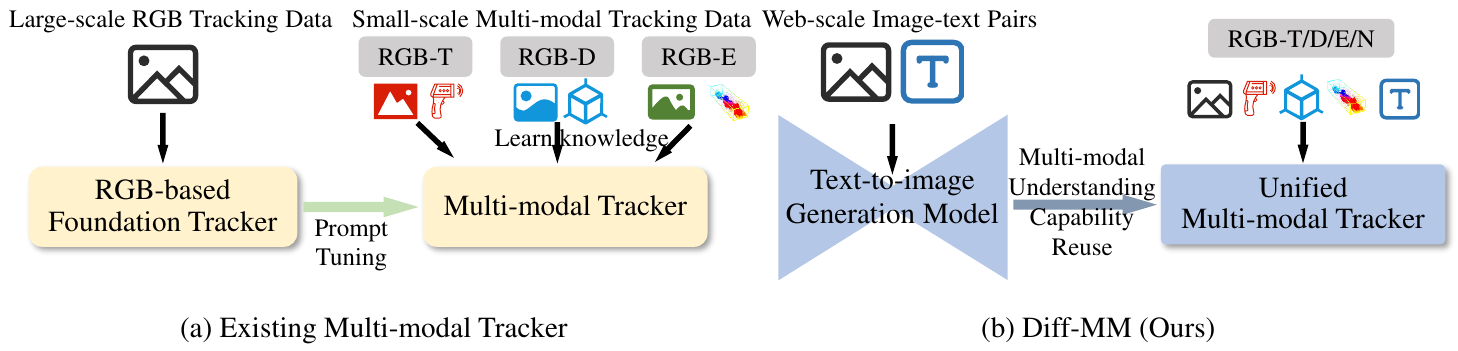}

   \caption{(a) Recent multi-modal trackers typically start from a foundation RGB tracker trained on large-scale RGB tracking data. They learn to understand auxiliary modality from small-scale multi-modal tracking data, using prompt-tuning or LoRA to avoid forgetting. (b) Our method acquires a unified multi-modal tracker by leveraging a text-to-image generation model. With the help of prior knowledge in the generation model, we achieve a unified tracker for RGB-N/D/T/E tracking.}
   \label{fig:motivation}
\end{figure}

As shown in Fig.~\ref{fig:motivation} (a), to achieve multi-modal tracking, recent methods~\cite{hong2024onetracker,zhu2023visual,wu2024single} typically follow a two-stage training pipeline, which trains a foundation RGB tracking model on large-scale RGB-based tracking datasets and fine-tunes the model through parameter-efficient tuning~\cite{vpt,hu2021lora} on downstream multi-modal tracking datasets.
These methods learn to understand auxiliary modalities solely from the training set. However, the available multi-modal tracking datasets are relatively small, \emph{e.g.}, DepthTrack~\cite{depthtrack} contains only 150 training sequences, which restricts the capability of these methods to understand auxiliary modalities.
Moreover, these methods only update a small number of parameters for auxiliary modality, making it difficult to achieve a generalist model that learns all modalities within a group of parameters.

A strong pre-trained model can significantly reduce the training data requirement.
Trained with billions of image-text pairs, the cutting-edge generation model (Stable Diffusion (SD)~\cite{rombach2022high}) shows impressive generation capability. More importantly, SD has been validated to have great potential in understanding modalities beyond RGB images, as it has been successfully transferred to tasks such as depth estimation~\cite{ke2024repurposing,fu2025geowizard}, conditional generation~\cite{zhang2023adding,mou2024t2i}, event data simulation~\cite{wu2025motion}, and real-to-thermal infrared translation~\cite{Mayr_2024_CVPR}. Given its ability to understand Text/Depth/Event flow/Thermal infrared, transferring the pre-trained SD to multi-modal tracking presents a promising avenue for overcoming dataset limitations and developing a unified multi-modal tracker. 

However, leveraging the SD for object tracking is not a trivial task. Firstly, unlike image generation, which takes a single image as input, object tracking requires a pair of frames, a template frame and a search frame. Secondly, SD cannot deal with modalities beyond RGB images directly. Modifying the model architecture or introducing extra modules is needed. This conflicts with reusing the prior knowledge in the pre-trained parameters.

To tackle this challenge, we propose a parallel feature extraction pipeline (PFE). PFE processes the template and search frames in parallel and performs relationship modeling between two features at the self-attention layer in the UNet. This design maintains the architecture of the UNet, better preserving the prior knowledge encoded in the parameters of the UNet. Meanwhile, it enables the reuse of text understanding capability of SD for RGB-N tracking by incorporating the language description of the object as a text condition.
To further integrate other auxiliary modalities, we introduce multi-modal sub-module tuning, which preserves the capability of the UNet by freezing its parameters while copying some modules from the UNet as the modality-specific sub-module.
%



With the help of extensive prior knowledge in SD, Diff-MM achieves a unified multi-modal tracker with a single set of parameters for RGB-N/D/T/E tracking.
Experimental results across several multi-modal tracking benchmarks, including RGB-T tracking (LasHeR~\cite{lasher}, RGBT234~\cite{rgbt234}), RGB-D tracking (DepthTrack~\cite{depthtrack}, VOT-RGBD2022~\cite{vot22}), RGB-E tracking (VisEvent~\cite{visevent}), and RGB-N tracking (TNL2K~\cite{wang2021towards}, OTB99~\cite{li2017tracking}), demonstrate promising performance of our method. It achieves a performance gain of 8.0\% on F-score over the recently proposed OneTracker~\cite{hong2024onetracker} on DepthTrack.

In summarize, the contributions of this work are:
\begin{itemize}[itemsep=0.1em,leftmargin=*,topsep=0em]
  \item Different from improving the architecture and head design for unified multi-modal tracking, this work explores a new way by incorporating the extensive knowledge encapsulated in SD.
  \item To transfer SD to multi-modal object tracking and maintain its knowledge priors, a parallel feature extraction pipeline and multi-modal sub-module tuning method is proposed.
  \item Diff-MM unifies RGB-N/D/T/E tracking under a single architecture with a single set of parameters and achieves superior performance across various benchmarks.
\end{itemize}

\section{Related Works}
\label{sec:related_work}
This work is closely related to visual object tracking and the use of diffusion models for visual perception. This section briefly reviews these related works.

\noindent\textbf{Multi-modal Tracking.} Single object tracking aims to predict the location and size of an object in a continuous video sequence based on an initial position given in the first frame~\cite{otb2015,vot18}. To deal with such a challenging task, the tracking framework has evolved from correlation filters~\cite{KCF,eco}, Siamese networks~\cite{siamesefc,siamrpnplusplus} to Transformer~\cite{stark,ostrack,transt}. Despite the promising performance, RGB-based trackers still face unstable tracking on low-illumination or fast-moving scenes due to the insufficient information provided by RGB data. To address this, multi-modal object tracking combines additional modalities to provide complementary information. For instance, in occlusion scenarios, RGB-D trackers can integrate depth information~\cite{depthtrack}. RGB-T trackers are robust to illumination variations~\cite{lasher,rgbt234}. Event flow enables RGB-E trackers to work well under fast motion and low illumination~\cite{visevent}. Combining visual data with natural language allows trackers to gain more semantic information about the tracked object~\cite{wang2021towards,zhou2023joint}. Despite these advancements, researchers still need to design specific modules for different modalities. Some efforts have been made to develop unified architectures that integrate multiple modalities, \emph{e.g.}, ViPT~\cite{zhu2023visual}, and Un-Track~\cite{wu2024single}. These methods follow a similar pipeline, where a foundation RGB tracker is augmented with additional modalities via prompt tuning~\cite{vpt} or LoRA~\cite{hu2021lora}. These methods focus on the architecture of the tracking features extractor. The limitation brought by the small-scale training data still cannot be solved.



\noindent\textbf{Diffusion Models for Visual Perception.} Diffusion models~\cite{ho2020denoising} are the next generation of image generation models after GANs~\cite{goodfellow2014generative}. Trained on billions of image-text pairs~\cite{schuhmann2022laion}, Stable Diffusion~\cite{rombach2022high} encodes rich image priors in its model weights, enabling it to generate diverse and high-quality images. Many efforts have explored the application of SD to visual perception tasks. Marigold~\cite{ke2024repurposing} and GeoWizard~\cite{fu2025geowizard} repurposes monocular depth estimation as an image generation task. DIFT~\cite{tang2023emergent} and VPD~\cite{zhao2023unleashing} use the pre-trained UNet in SD as a feature extractor for visual correspondence and dense prediction tasks, respectively.
Diff-Tracker~\cite{zhang2024diff} learns target prompt tokens for unsupervised object tracking. Those efforts demonstrate the potential of the pre-trained SD for visual perception tasks.

\noindent\textbf{Differences with Previous Works.}
Existing unified multi-modal trackers focus on the architecture of the feature extractor to better integrate information from auxiliary modalities, whose performance is limited by the small training set of multi-modal tracking. In contrast, our method alleviates the issue of insufficient multi-modal tracking data by incorporating the extensive knowledge encapsulated in SD, which has been verified to exhibit capabilities in understanding many modalities, \emph{e.g.}, depth~\cite{ke2024repurposing}, event flow~\cite{wu2025motion}, and thermal infrared~\cite{Mayr_2024_CVPR}. Both Diff-Tracker and our method harness SD for object tracking. However, Diff-Tracker focuses on unsupervised RGB-based tracking and requires time-consuming online learning for each object. As demonstrated in Sec.~\ref{sec:ablation}, this learning method does not perform well in supervised object tracking and is difficult to extend to multi-modal tracking. Our method does not rely on online learning and achieves unified multi-modal tracking across RGB-N/D/T/E, all under a single model architecture with uniform parameters.

\section{Methodology}
\label{sec:method}

\begin{figure*}[h!]
    \centering
     \includegraphics[width=0.95\linewidth]{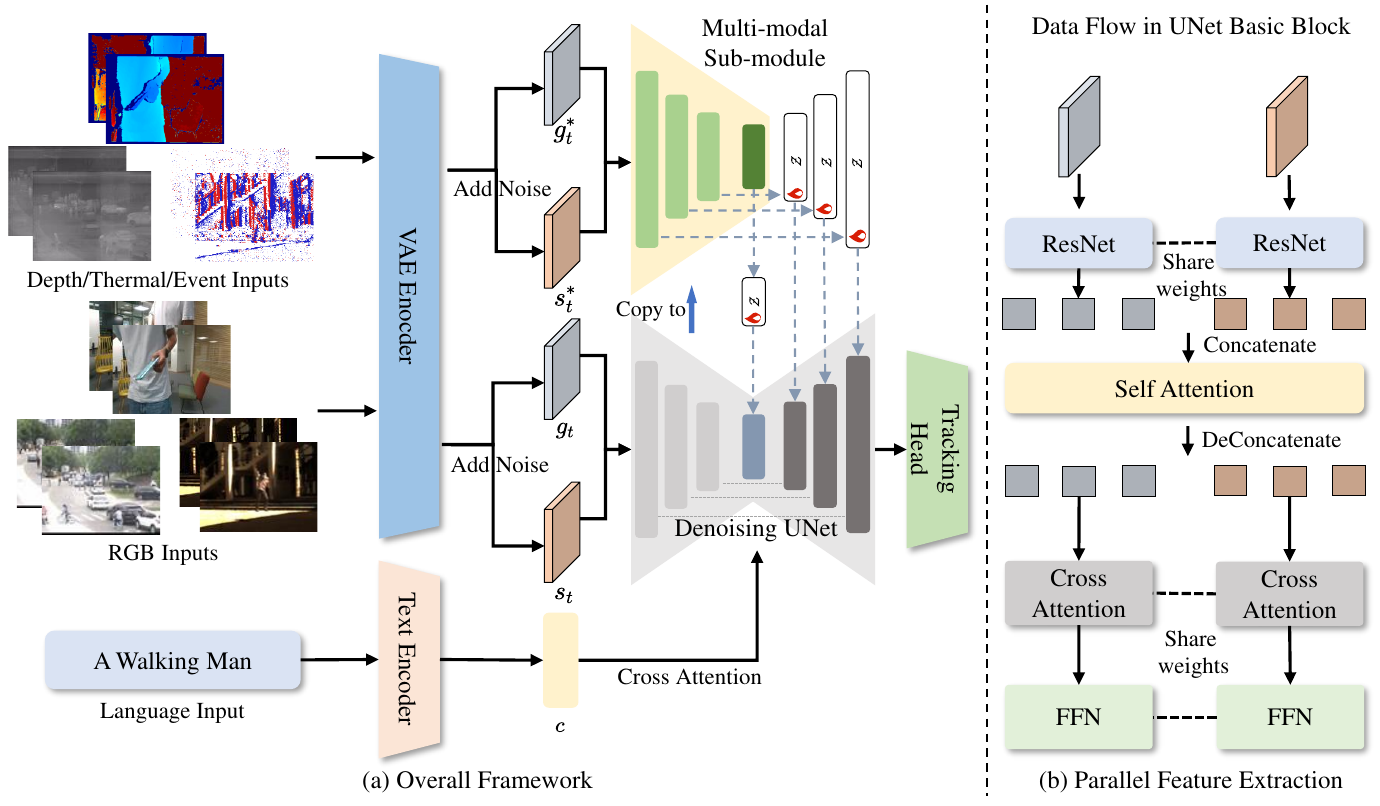}
  
     \caption{Illustrations of the overall framework and the parallel feature extraction pipeline of Diff-MM. (a) We achieve a unified model for RGB-N/D/T/E tracking by exploiting generative priors from the pre-trained SD. For RGB-D/T/E tracking, we first encoder the RGB/Depth/Thermal/Event images into the latent space with the VAE encoder and noise adding. Latent features of RGB images are sent into the UNet. We introduce a multi-modal sub-module copied from the encoder and middle block of the UNet to model auxiliary modalities. Lateral-connection is used to inject auxiliary information into the UNet. The output features are fed into the tracking head for bounding box prediction. The language description of the object is used as the text condition to perform RGB-N tracking. (b) The parallel feature extraction pipeline enables pairwise image inputs by extracting template features and search features in parallel and modeling their relationship at the self attention layer in each Basic Block of the UNet.}
     \label{fig:overall}
  \end{figure*}

\subsection{Overview}
Given an initial bounding box in the first frame, object tracking aims to locate this object in a video sequence. Typically, given a template frame $T$ contains the information of tracked object, and a search frame $S$, the tracker extracts the tracking features $f = \Phi(T, S)$ with a feature extractor $\Phi$ to model the relationship between these two frames. The tracking features are subsequently fed into the tracking head for bounding box prediction $B=\text{Head}(f)$. To address the limitation of RGB-based tracking, multi-modal tracker further integrates auxiliary modalities into the tracking features extraction pipeline,
\begin{equation}
    \begin{aligned}
        f &= \Phi(T, S, T^*, S^*), \text{RGB-D/T/E}, \\
        f &= \Phi(T, S, N), \text{RGB-N},
    \end{aligned}\label{eq:feature_extraction}
\end{equation} 
where $T^*$ and $S^*$ denote the template and search frames of auxiliary modality in RGB-D/T/E tracking, and $N$ denotes the text description of the tracked object in RGB-N tracking.

$\Phi(\cdot)$ models the relationship between two frames, which is crucial in object tracking. Based on $\Phi(\cdot)$ trained on large-scale RGB-based tracking datasets, typical multi-modal trackers leverage downstream multi-modal tracking datasets to equip $\Phi(\cdot)$ with multi-modal understanding ability. Recent researches focus on the improvement of $\Phi(\cdot)$ structure to better integrate information from different modalities~\cite{zhu2023visual,wu2024single,tan2024towards}. However, the capability of $\Phi(\cdot)$ is still limited by the data scale of multi-modal tracking datasets.

A strong pre-trained model can significantly reduce the training data requirement. Therefore, we exploit the rich knowledge in the text-to-image model (Stable Diffusion~\cite{rombach2022high}) trained with billions of image-text pairs. SD has promising ability in understanding modalities beyond RGB images, as it has been successfully transferred to multi-modal tasks such as depth estimation~\cite{ke2024repurposing}, event data simulation~\cite{wu2025motion}, and real-to-thermal infrared translation~\cite{Mayr_2024_CVPR}. This motivates us to learn a unified tracking feature extractor $\Phi(\cdot)$ for RGB-N/T/D/E tracking based on the pre-trained weights of SD.
SD contains a VAE encoder $z_0=\mathcal{E}(Z)$ and decoder $Z = \mathcal{D}(z_0)$ that translate an image $Z$ into/back from latent spaces, and a denoising UNet model $\epsilon_{\theta}(\cdot)$ that takes the noisy latent features, text condition $c$ and timestep $t$ as input and predicts the added noise $\epsilon=\epsilon_{\theta}(z_t, c, t)$.

To reuse the knowledge prior in SD for visual perception tasks, VPD~\cite{zhao2023unleashing} and DIFT~\cite{tang2023emergent} utilize the pre-trained UNet as a feature extractor by removing its noise prediction module. Maintaining the architecture of the UNet is important to the reuse of the knowledge prior. However, the UNet cannot be directly used as a tracking feature extractor. Object tracking requires a pair of frames as input, a template frame $T$ and a search frame $S$, and needs to model their relationship in the feature extraction pipeline. To handle this pairwise input and minimize the modification of the UNet, we propose a parallel feature extraction pipeline (PFE), which is conceptually formulated as,
\begin{equation}
    f = \text{PFE}(T, S, c;\mathcal{E}, \epsilon_{\theta}),
    \label{eq:PFE}
\end{equation}
where $f$ denotes the output tracking features. 

PFE enables the UNet to be a tracking feature extractor and also leads to the reuse of SD's text understanding capability for RGB-N tracking by incorporating the language description of the object as the text condition.
To further integrate the feature extractor with other modalities (D, E, T), inspired by the conditional control image generation methods~\cite{zhang2023adding,mou2024t2i}, we introduce a multi-modal sub-module tuning method (MST). This method copies some modules from the UNet to construct a sub-module for auxiliary modalities, allowing the integration of prior knowledge from the UNet to handle diverse modalities. Through PFE and MST, we develop a unified model for multi-modal object tracking as shown in Fig.~\ref{fig:overall}.
The detailed implementation of PFE and MST will be presented in the following parts.

\subsection{Parallel Feature Extraction}
Object tracking requires two inputs: a template frame $T$ and a search region frame $S$.
To change the pre-trained UNet of SD as a tracking feature extractor, an interaction mechanism is needed to model the relationship between the template and search frames.
One straightforward approach is to treat the template frame as a condition and inject object information through cross attention. However, this design would introduce an additional module to extract image features of the template frame, such as a ViT~\cite{vit}, which cannot benefit from the prior knowledge in SD. Differently, Diff-Tracker~\cite{zhang2024diff} introduces online learning to embed object information into object prompt tokens. Learning object prompt tokens for each object is time-consuming and insufficient for capturing fine-grained details~\cite{zhu2024exploring}.

To reuse the architecture and parameters of SD without introducing additional modules, we propose a parallel feature extraction pipeline (PFE).
PFE treats the template frame and the search frame symmetrically. Specifically, these two frames are first mapped into the latent space with the VAE encoder $s_0=\mathcal{E}(S)$, $g_0=\mathcal{E}(T)$, separately. To avoid disturbing the training distribution of the UNet, the diffusion forward process is applied to generate the noisy latent features $s_t$ and $g_t$. These noisy features are then passed through the UNet to obtain the final tracking features.

The UNet consists of an encoder, a middle block, and a decoder. These modules are constructed by stacking Basic Blocks with additional down/up sampling convolution layers. As shown in Fig.~\ref{fig:overall} (b), Basic Block is built with several ResNet blocks and Transformer blocks. To maintain the architecture of Basic Block, PFE processes the search and template features separately and performs the interaction between two features at the self-attention layer. The data flow of PFE in Basic Block can be written as,
\begin{equation}
        \small
    \begin{aligned}
    s_t[j], g_t[j] &\leftarrow \text{ResNet}_j(s_t[j],t), \text{ResNet}_j(g_t[j],t), \\
    O &\leftarrow \text{ConcatL}(s_t[j], g_t[j]), \\
    O &\leftarrow \text{SA}_j(O), \\
    s_t[j], g_t[j] &\leftarrow \text{DeConcatL}(O), \\
    s_t[j], g_t[j] &\leftarrow \text{CA}_j(s_t[j], c), \text{CA}_j(g_t[j], c),\\
    s_t[j+1], g_t[j+1] &\leftarrow \text{FFN}_j(s_t[j]), \text{FFN}_j(g_t[j]),
    \end{aligned}\label{eq:PFE_detail}
\end{equation}
where $j$ denotes the $j$-th Basic Block. $\text{ResNet}_j$, $\text{SA}_j$, $\text{CA}_j$, and $\text{FFN}_j$ represent the ResNet, self attention layer, cross attention layer, and FFN, respectively. ConcatL and DeConcatL are the concatenation and de-concatenation operations along the length dimension of the vectors. For simplicity, we omit Layer/Group Normalization and down/up sampling convolution layers.


PFE avoids modifying the architecture and introducing additional modules, which is key to preserving the prior knowledge of the SD. It also enables the reuse of SD's text understanding capability and achieves RGB-N tracking by using the language description of the object as the text condition $c$.
For datasets that do not have language descriptions, we simply use a \textit{null string} as the text condition.

  

\subsection{Multi-modal Sub-module Tuning}\label{sec:MST}
To develop a unified model for multi-modal tracking, beyond the language description in RGB-N tracking, the feature extractor should be capable of integrating complementary information from auxiliary modalities such as depth, thermal infrared, and event flow. Multi-modal sub-module tuning (MST) achieves this by incorporating these auxiliary modalities with a sub-module, which reuses the parameters of the UNet and follows the proposed PFE.

Given the auxiliary modality of template $T^*$ and search $S^*$ frames, we apply the VAE encoding stage followed by a forward diffusion process to obtain their noisy latent features $g^*_t$ and $s^*_t$. Since RGB and auxiliary modalities have different data distributions, it's essential to retain the model's capability for RGB representation while simultaneously learning to process auxiliary modalities. As shown in Fig.~\ref{fig:overall}, we freeze the parameters of the UNet, and construct a sub-module by cloning the parameters of encoder and middle block. $g^*_t$ and $s^*_t$ are sent into the cloned encoder and middle block to extract auxiliary features, encapsulating information of the auxiliary modality. These auxiliary features are merged with the RGB features from the middle block or with the RGB lateral-connection features from the corresponding encoder layer,
\begin{equation}
    \begin{aligned}
    \hat{s}_t[N-i] &\leftarrow \hat{s}_t[N-i] + \mathcal{Z}(\hat{s}^*_t[N-i]), \\
    \hat{g}_t[N-i] &\leftarrow \hat{g}_t[N-i] + \mathcal{Z}(\hat{g}^*_t[N-i]),
    \end{aligned}\label{eq:MST}
\end{equation}
where $\hat{s}_t[N-i]$ denotes the lateral-connection features, $\mathcal{Z}$ denotes a zero-init convolution layer. $\mathcal{Z}$ ensures that the auxiliary modality does not influence the trained RGB network at the beginning of the training.

\subsection{Training and Inference}\label{sec:training_inference}
\noindent\textbf{Training.} Following the same pipeline of most multi-modal trackers~\cite{hong2024onetracker,protrack,zhu2023visual}, the model is trained with two stages. In the first stage, we fine-tune the self attention layer of UNet and the tracking head on large-scale RGB and RGB-N tracking datasets. In the second stage, as described in Sec.~\ref{sec:MST}, we clone the encoder and the middle block trained in the first stage to create the sub-module. We freeze the UNet and tracking head, and tune the sub-module on RGB-D/T/E tracking datasets. In addition to the modal-specific fine-tuning, due to the powerful capacity of the UNet, we can achieve a generalist sub-module with only one parameter set by tuning the model on the merged multi-modal tracking datasets.

The tracking head is similar to OSTrack~\cite{ostrack}. Features of the search frame are used as the final tracking features. The overall training loss is the same in the two training stages and is given,
\begin{equation}
    \mathcal{L} = L_{cls} + \lambda_{giou} L_{giou} + \lambda_{L_1} L_1, \\
    \label{eq:overall}
\end{equation}
where $L_{cls}$ is the focal loss~\cite{law2018cornernet} for classification, $L_{giou}$ and $L_1$ denote the GIOU loss~\cite{rezatofighi2019generalized} and L1 loss for bounding box regression, $\lambda_{giou}$ and $\lambda_{L_1}$ are the loss weights.

\noindent\textbf{Inference.} For RGB-N tracking, the model without the multi-modal sub-module is adopted. For RGB-D/T/E tracking, the tracking can be performed by adding the sub-module into the model.
Similar to other multi-modal trackers, we transform the depth map, thermal-infrared image, and event flow into an RGB-like form.
Due to the specific design of our method, we can achieve multi-modal tracking with a unified model architecture and parameter set.

\section{Experiments}\label{sec:experiment}
\subsection{Experimental Settings}
\noindent\textbf{Training Data.} Our method follows a two-stage training pipeline. In the first stage, a combination of several large-scale RGB and RGB-N tracking datasets is used, including GOT-10K~\cite{got10k}, LaSOT~\cite{lasot}, TrackingNet~\cite{trackingnet}, TNL2K~\cite{wang2021towards} and COCO~\cite{coco}. In the second stage, we choose training set of DepthTrack~\cite{depthtrack}, LasHeR~\cite{lasher}, VisEvent~\cite{visevent} for RGB-D/T/E tracking, respectively.




\noindent\textbf{Implementation Details.} We utilize Stable Diffusion v1-5~\cite{rombach2022high} as the base model. The VAE encoder is frozen during training. The timestep for the diffusion forward process is 1. In the first training stage, we train the model using AdamW optimizer~\cite{loshchilov2017decoupled} with a total batch size 128. The weight decay is set to $10^{-3}$. The learning rate for the UNet and the tracking head is set to $4 \times 10^{-6}$ and $4 \times 10^{-4}$, respectively, and decays according to the cosine annealing rule. The model is trained for 300 epochs with 60k image pairs per epoch. In the second training stage, for RGB-D/T/E specific tracking, we copy the parameters of the encoder and middle block from the UNet trained in the first stage to form the sub-module and perform multi-modal sub-module tuning. The UNet and tracking head are frozen. For the generalist multi-modal tracking, we train only one sub-module for all modalities with the combination of RGB-D/T/E tracking training sets. The model is trained for 60 epochs, with an initial learning rate of $4 \times 10^{-6}$. Following OSTrack~\cite{ostrack}, $\lambda_{giou}$ and $\lambda_{L_1}$ are set to 2 and 5, respectively.
We use ZeRO-2~\cite{rajbhandari2020zero} offload to reduce the memory requirements, which allows us to train the model on 4 NVIDIA RTX 3090 GPUs.
For a fair comparison, we present two variants of our tracker: Diff-MM 256 and Diff-MM 384, corresponding to search region sizes of 256 and 384, respectively. Moreover, to show the effectiveness of our design to other diffusion architectures, we also present Diff-MM 256$\dagger$, which uses SD v3-5 as the base model. SD v3-5 uses MM-DiT~\cite{peebles2023scalable} instead of the UNet.

\subsection{Main Results}
\begin{table*}
    \caption{Overall performance on RGB-N/D/T/E tracking. ``Diff-MM'' is our method. ``Diff-MM$\dagger$'' denotes our method with SD v3-5 as the base model.}
    \label{tab:sota}
    \centering
    \setlength{\tabcolsep}{0.8px}
    \scriptsize
     \resizebox{\linewidth}{!}{
    \begin{tabular}{cc|cc|cccc|cc|cc|ccc|ccc}
    \toprule[2pt]
    
    & \multicolumn{17}{c}{\textbf{RGB-N Tracking}} \\ \midrule
    &&\multicolumn{2}{c|}{\textbf{Modality-specific Model}} & \multicolumn{8}{c|}{\textbf{Unified Model with Modality-specific Parameters}} & \multicolumn{6}{c}{\textbf{Unified Model with Single Set of Parameters}}\\
    \midrule
    & &\tabincell{c}{SNLT\\~\cite{feng2021siamese}} &\tabincell{c}{JointNLT\\~\cite{zhou2023joint}} & \tabincell{c}{ProTrack\\~\cite{protrack}} &\tabincell{c}{ViPT\\~\cite{zhu2023visual}} &\tabincell{c}{Un-Track\\~\cite{wu2024single}} & \tabincell{c}{SDSTrack\\~\cite{hou2024sdstrack}} &\tabincell{c}{\textbf{Diff-MM}\\\textbf{256}} & \tabincell{c}{\textbf{Diff-MM}\\\textbf{256}$\dagger$} & \tabincell{c}{OneTracker\\384~\cite{hong2024onetracker}} & \tabincell{c}{\textbf{Diff-MM}\\\textbf{384}} & \tabincell{c}{ViPT\\~\cite{zhu2023visual}} &\tabincell{c}{Un-Track\\~\cite{wu2024single}} &\tabincell{c}{XTrack\\~\cite{tan2024towards}} & \tabincell{c}{\textbf{Diff-MM}\\\textbf{256}} & \tabincell{c}{\textbf{Diff-MM}\\\textbf{256}$\dagger$} & \tabincell{c}{\textbf{Diff-MM}\\\textbf{384}} \\
    
    \midrule
    \multirow{2}{*}{\tabincell{c}{OTB99\\~\cite{li2017tracking}}}&AUC($\uparrow$)  &66.6 &65.3 & - & - &- & -  & \textbf{71.9}& \textbf{72.9} & 69.7& \textbf{73.8} &- &- & - & \textbf{71.9} &\textbf{72.9} &\textbf{73.8} \\
    &P($\uparrow$)  &80.4 &85.6 & - & - & - & - & \textbf{93.5} & \textbf{94.3} &91.5  & \textbf{95.2} &- &- & - & \textbf{93.5} &\textbf{94.3} & \textbf{95.2} \\
    \midrule
    \multirow{2}{*}{\tabincell{c}{TNL2K\\~\cite{wang2021towards}}}&AUC($\uparrow$)&27.6 &56.9 & - & - &- & - & \textbf{65.2}& \textbf{67.5}& 58.0  & \textbf{66.3}&- &- & - & \textbf{65.2} &\textbf{67.5}&\textbf{66.3}\\
    &P($\uparrow$) &41.9 &58.1 & - & - &- & -& \textbf{70.1} & \textbf{72.9} & 59.1  & \textbf{72.3} &- &- & - & \textbf{70.1} & \textbf{72.9} &\textbf{72.3} \\

    \midrule[2pt]

    & \multicolumn{17}{c}{\textbf{RGB-D Tracking}} \\ \midrule
    &&\multicolumn{2}{c|}{\textbf{Modality-specific Model}} & \multicolumn{8}{c|}{\textbf{Unified Model with Modality-specific Parameters}} & \multicolumn{6}{c}{\textbf{Unified Model with a Single Set of Parameters}}\\
    \midrule
    &&\tabincell{c}{OSTrack\\~\cite{ostrack}}&\tabincell{c}{SPT\\~\cite{rgbd1k}}&\tabincell{c}{ProTrack\\~\cite{protrack}} &\tabincell{c}{ViPT\\~\cite{zhu2023visual}} &\tabincell{c}{Un-Track\\~\cite{wu2024single}} & \tabincell{c}{SDSTrack\\~\cite{hou2024sdstrack}}  &\tabincell{c}{\textbf{Diff-MM}\\\textbf{256}} & \tabincell{c}{\textbf{Diff-MM}\\\textbf{256}$\dagger$} & \tabincell{c}{OneTracker\\384~\cite{hong2024onetracker}} & \tabincell{c}{\textbf{Diff-MM}\\\textbf{384}} & \tabincell{c}{ViPT\\~\cite{zhu2023visual}} &\tabincell{c}{Un-Track\\~\cite{wu2024single}} &\tabincell{c}{XTrack\\~\cite{tan2024towards}} & \tabincell{c}{\textbf{Diff-MM}\\\textbf{256}} & \tabincell{c}{\textbf{Diff-MM}\\\textbf{256}$\dagger$} & \tabincell{c}{\textbf{Diff-MM}\\\textbf{384}} \\

    \midrule
    \multirow{3}{*}{\tabincell{c}{DepthTrack\\~\cite{depthtrack}}}&F-score($\uparrow$)&52.9&53.8&57.8 &59.4 & 61.2 & 61.4 & \textbf{65.3}& \textbf{67.4}& 60.7 & \textbf{68.7} &56.1 & 61.0 & 59.7 & \textbf{65.9} &\textbf{67.6}& \textbf{68.4}\\
    &Re($\uparrow$)&52.2&54.9&57.3&59.6 & 61.0 & 60.9 & \textbf{65.0}& \textbf{67.3}& 60.9  &\textbf{68.7} & 56.2 & 61.0 & 59.7 & \textbf{65.5} &\textbf{67.6}& \textbf{68.5}\\
    &Pr($\uparrow$)&53.6&52.7&58.3&59.2 & 61.3 & 61.9 & \textbf{65.6}&\textbf{67.5}& 60.4  &\textbf{68.7} & 56.0 & 61.0 & 59.8 & \textbf{66.2} & \textbf{67.6}&\textbf{68.3}\\
    \midrule
    \multirow{3}{*}{\tabincell{c}{VOT\\RGBD2022\\~\cite{vot22}}}& EAO($\uparrow$)&67.6&65.1&65.1&72.1& - & 72.8 & \textbf{76.4}& \textbf{78.4}& 72.7  & \textbf{78.0} &- & 71.8 & 71.4 & \textbf{77.2} &\textbf{78.6}& \textbf{78.3} \\
    &Accuracy($\uparrow$)&80.3&79.8&80.1&81.5 & -& 81.2 & \textbf{81.5} & \textbf{82.5} & 81.9  & \textbf{82.2} &- & 82.0 & 81.2 & \textbf{82.0} & \textbf{82.7} & \textbf{82.4}\\
    &Robustness($\uparrow$)&83.3&85.1&80.2&87.1 & -& 88.3 & \textbf{93.0} & \textbf{94.4} & 87.2  & \textbf{93.9} &- & 86.4 & 86.5 & \textbf{93.7} &\textbf{94.4} & \textbf{94.2}\\

    \midrule[2pt]

    & \multicolumn{17}{c}{\textbf{RGB-T Tracking}} \\ \midrule
    &&\multicolumn{2}{c|}{\textbf{Modality-specific Model}} & \multicolumn{8}{c|}{\textbf{Unified Model with Modality-specific Parameters}} & \multicolumn{6}{c}{\textbf{Unified Model with a Single Set of Parameters}}\\
    \midrule
    & &\tabincell{c}{DAFNet\\~\cite{dafnet}} &\tabincell{c}{STMT\\~\cite{10589660}}&\tabincell{c}{ProTrack\\~\cite{protrack}} &\tabincell{c}{ViPT\\~\cite{zhu2023visual}} &\tabincell{c}{Un-Track\\~\cite{wu2024single}} & \tabincell{c}{SDSTrack\\~\cite{hou2024sdstrack}}  &\tabincell{c}{\textbf{Diff-MM}\\\textbf{256}}& \tabincell{c}{\textbf{Diff-MM}\\\textbf{256}$\dagger$} & \tabincell{c}{OneTracker\\384~\cite{hong2024onetracker}} & \tabincell{c}{\textbf{Diff-MM}\\\textbf{384}} & \tabincell{c}{ViPT\\~\cite{zhu2023visual}} &\tabincell{c}{Un-Track\\~\cite{wu2024single}} &\tabincell{c}{XTrack\\~\cite{tan2024towards}} & \tabincell{c}{\textbf{Diff-MM}\\\textbf{256}} & \tabincell{c}{\textbf{Diff-MM}\\\textbf{256}$\dagger$} & \tabincell{c}{\textbf{Diff-MM}\\\textbf{384}} \\
    \midrule
    \multirow{2}{*}{\tabincell{c}{LasHeR\\~\cite{lasher}}} &SR($\uparrow$) &31.1 & 53.7 &42.0 &52.5 & 53.6 & 53.1 & \textbf{57.0}& \textbf{62.4}& 53.8 &\textbf{57.9} & 49.0 & 51.3 & 52.5 & \textbf{57.5} & \textbf{62.7}& \textbf{58.2}\\
    &PR($\uparrow$) & 44.8 & 67.4 &53.8 &65.1  & 66.7 & 66.5 & \textbf{71.3}& \textbf{77.9}&67.2 & \textbf{72.6} & 60.8 & 64.6 & 65.5 & \textbf{72.1} & \textbf{78.0}& \textbf{73.0}\\
    \midrule
    \multirow{2}{*}{\tabincell{c}{RGBT234\\~\cite{rgbt234}}}&MSR($\uparrow$) & 54.4 & 63.8 &59.9 &61.7 & 61.8 & 62.5 & \textbf{69.1}& \textbf{71.2}& 64.2 &\textbf{70.0} & - & 62.5 & 62.2 & \textbf{69.1} &\textbf{71.1}& \textbf{70.2}\\
    &MPR($\uparrow$) & 79.6 & 86.5 &79.5 &83.5 & 83.7 & 84.8 & \textbf{91.1}& \textbf{94.8}& 85.7 & \textbf{92.2} & - & 84.2 & 84.8 & \textbf{91.3} & \textbf{94.8}&\textbf{92.7}\\

    \midrule[2pt]

    & \multicolumn{15}{c}{\textbf{RGB-E Tracking}} \\ \midrule
    &&\multicolumn{2}{c|}{\textbf{Modality-specific Model}} & \multicolumn{7}{c|}{\textbf{Unified Model with Modality-specific Parameters}} & \multicolumn{5}{c}{\textbf{Unified Model with a Single Set of Parameters}}\\
    \midrule
    &&\tabincell{c}{MDNet\\~\cite{mdnet}} &\tabincell{c}{PrDiMP\_E\\~\cite{prdimp}}&\tabincell{c}{ProTrack\\~\cite{protrack}} &\tabincell{c}{ViPT\\~\cite{zhu2023visual}} &\tabincell{c}{Un-Track\\~\cite{wu2024single}} & \tabincell{c}{SDSTrack\\~\cite{hou2024sdstrack}} & \tabincell{c}{\textbf{Diff-MM}\\\textbf{256}}& \tabincell{c}{\textbf{Diff-MM}\\\textbf{256}$\dagger$} & \tabincell{c}{OneTracker\\384~\cite{hong2024onetracker}} & \tabincell{c}{\textbf{Diff-MM}\\\textbf{384}} & \tabincell{c}{ViPT\\~\cite{zhu2023visual}} &\tabincell{c}{Un-Track\\~\cite{wu2024single}} &\tabincell{c}{XTrack\\~\cite{tan2024towards}} & \tabincell{c}{\textbf{Diff-MM}\\\textbf{256}}& \tabincell{c}{\textbf{Diff-MM}\\\textbf{256}$\dagger$} & \tabincell{c}{\textbf{Diff-MM}\\\textbf{384}} \\
    \midrule
    \multirow{2}{*}{\tabincell{c}{VisEvent\\~\cite{visevent}}}&SR($\uparrow$) & 42.6 &45.3 & 47.1 & 59.2 & 59.7 & 59.7 & \textbf{61.9}& \textbf{64.8}& 60.8 &\textbf{63.3} & 57.9 & 58.9 & 59.1 & \textbf{61.9} & \textbf{64.6} & \textbf{63.2}\\
    &PR($\uparrow$) &66.1 &64.4 & 63.2 & 75.8 & 76.3 & 76.7  & \textbf{78.4}& \textbf{81.5}& 76.7 & \textbf{79.9} & 74.0 & 75.5 & 75.6 & \textbf{78.7} & \textbf{81.7}&\textbf{80.0}\\

    \bottomrule[2pt]
    
\end{tabular}

}
\end{table*}

To validate the effectiveness of our method, we conduct experiments on a wide range of multi-modal tracking benchmarks. The compared methods can be divided into three groups: 1) \textbf{Modality-specific trackers} which are designed for a specific RGB-X tracking, \emph{e.g.}, JointNLT~\cite{zhou2023joint} for RGB-N tracking, and DeT~\cite{depthtrack} for RGB-D tracking. 2) \textbf{Unified model with modality-specific parameters} which uses a unified model architecture but separate parameters for each modality, \emph{e.g.} ProTrack~\cite{protrack} and UnTrack~\cite{wu2024single}. 3) \textbf{Unified model with a single set of parameters}, \emph{e.g.}, XTrack~\cite{tan2024towards}. In addition, OneTracker~\cite{hong2024onetracker} uses high resolution inputs. To make a fair comparison, we compare with it using the high resolution version of our method. The comparison results are shown in Table~\ref{tab:sota}.

\noindent\textbf{RGB-N Tracking.} We conduct experiments on OTB99 and TNL2K. OTB99 is an extended version of OTB~\cite{otb2015} where each object is annotated with a language description. TNL2K is a large-scale benchmark containing 700 sequences for testing. The evaluation metrics include Area Under Curve (AUC) and Precision (P).
As shown in Table~\ref{tab:sota}, most generalist trackers cannot handle RGB-N tracking. Although OneTracker achieves a generalist model, the modality-specific parameters are still needed. Compared with other methods that need to align visual and text modalities from scratch, our design enables reuse of the SD's text understanding capability, leading to a superior performance, \emph{e.g.}, we outperform JointNLT by 8.3\% in AUC on TNL2K.

\noindent\textbf{RGB-D Tracking.} We evaluate our method on DepthTrack and VOT-RGBD2022. The DepthTrack testing set consists of 50 sequences, with Precision (Pr), Recall (Re), and F-score as evaluation metrics. VOT-RGBD2022 is constructed with 127 video sequences, and the evaluation metrics are Accuracy, Robustness and Expected Average Overlap (EAO). Similar to RGB-N tracking, our method also surpasses all compared trackers. The F-score of our method on DepthTrack is 65.9\%, outperforming previous best performance tracker Un-Track by 4.9\%. These results demonstrate the effectiveness of our method for RGB-D tracking.

\noindent\textbf{RGB-T Tracking.} LasHeR is the largest RGB-T tracking dataset with 245 test sequences. The Precision Rate (PR) and Success Plots (SR) are used as the metrics. RGBT234 includes 234 test sequences. Different from LasHeR, where the RGB and thermal images are well aligned, the two modalities in RGBT234 have slight alignment errors. Therefore, the Maximum Precision Rate (MPR) and Maximum Success Plots (MSR) are defined to measure the tracker performance. The experimental results demonstrate that our method excels in RGB-T tracking.

\noindent\textbf{RGB-E Tracking.} VisEvent consists of 320 test sequences. In each frame, RGB images and event flow are included. The evaluation metrics are Precision Rate (PR) and Success Rate (SR). With the modality-specific parameters, our method achieves a 2.2\% absolute gain in SR compared to Un-Track. Moreover, our model with a single set of parameters surpasses the XTrack by about 2.8\% in SR.

\noindent\textbf{Summary.} Our method achieves a unified model for multi-modal tracking and shows promising performance across RGB-N/D/T/E tracking. The generalist trackers only update a small set of parameters for multi-modal tracking to prevent forgetting the learned knowledge of RGB-based foundation trackers. This design limits the model capacity and makes these trackers cannot benefit from jointly training across multi modalities datasets, leading to a performance degradation, \emph{e.g.}, the SR of ViPT drops from 52.5\% to 49.0\% on LasHeR.
In contrast, our method leverages the powerful SD. With its high model capacity, our method achieves comparable performance both with a single set of parameters and with modality-specific parameters. These results highlight the potential of leveraging prior knowledge from SD. In addition, with SD v3-5 as the base model, our method also shows promising performance, demonstrating the effectiveness of our design to other architectures.

\subsection{Ablation Studies}\label{sec:ablation}

\noindent\textbf{Effectiveness of PFE.} PFE is a key component of our method that enables the pre-trained UNet to process pairwise image inputs. To validate that PFE is a reasonable choice, we evaluate the model trained after the first training stage on two widely used RGB tracking benchmarks: LaSOT~\cite{lasot} and TrackingNet~\cite{trackingnet}. We report the Area under Curve (AUC), Normalized Precision ($\text{P}_{Norm}$), and Precision (P). The results are summarized in Table~\ref{tab:pfe} \textbf{Left}. For comparison, we re-implement Diff-Tracker~\cite{zhang2024diff}, a tracking method that also leverages generative models. It learns the object prompt tokens online to encoder object information for unsupervised tracking. Despite re-training Diff-Tracker with ground-truth annotations, the results remain unsatisfactory, achieving only 58.7\% AUC on LaSOT. This indicates that encoding object information into prompt tokens lacks the fine-grained detail necessary for object tracking.

Reference-based image in-painting~\cite{zhang2023paste} and SD image variations~\cite{sd-image-variations} inject the reference image information through the cross attention with an extra image encoder. We conduct experiments with three different settings. While these methods show performance improvements over Diff-Tracker, they are still outperformed by PFE. These methods introduce an extra image encoder which does not fully utilize the knowledge in SD. In addition, these designs cannot support RGB-N tracking. These results demonstrate that it is important to minimize the modification to the architecture of the SD.

\begin{table}
    \caption{\textbf{Left:} Ablation studies on PFE. ``Diff-Tracker'' embeds object information into the object prompt tokens. ``Cross-Attn'' refers to adopting the template image as a condition and injecting object information through cross-attention. ``MAE'' and ``CLIP ViT'' indicate the use of corresponding backbone for extracting template features. $\dagger$ denotes freezing the CLIP ViT. \textbf{Right:} Ablation studies on MST. ``RGB Only'' refers to the single-modal baseline. ``Cross-Attn'' denotes adopting the auxiliary modality image as a condition. ``Early Fusion'' indicates fusing RGB and auxiliary modality latent features at the beginning of the UNet. ``Extra Encoder'' introduces an extra small modality encoder as in T2I-Adapter~\cite{mou2024t2i}. ``w/o Zero Init'' denotes removing the zero-init convolution layer. ``Tuning UNet'' denotes tuning the UNet during the second training stage. * means the performance of RGB-N tracking is influenced. $\dagger$ denotes TNL2K is not used at the first stage.}
    \label{tab:pfe}
    \begin{minipage}[c]{0.5\linewidth}
    \centering
    \setlength{\tabcolsep}{0.8px}
    \small
     \resizebox{\linewidth}{!}{
    \begin{tabular}{c|ccc|ccc|c}
    \toprule[2pt]
    \multirow{2}{*}{Settings} & \multicolumn{3}{c|}{LaSOT} & \multicolumn{3}{c|}{TrackingNet} & \multirow{2}{*}{\tabincell{c}{RGB-N \\Tracking}} \\ \cline{2-7}
    & AUC & $\text{P}_{Norm}$ & P & AUC & $\text{P}_{Norm}$ & P & \\
    \midrule
    \tabincell{c}{Diff-Tracker} & 58.7 & 67.3 & 58.2 & 75.6 & 80.8 & 70.3 & \emph{No} \\
    \tabincell{c}{Cross-Attn +\\MAE} & 65.7 & 75.0 & 70.7 & 81.2 & 86.7 & 79.6 & \emph{No} \\
    \tabincell{c}{Cross-Attn +\\CLIP ViT} & 64.5 & 73.7 & 68.9 & 81.1 & 85.9 & 79.2 & \emph{No} \\
    \tabincell{c}{Cross-Attn +\\CLIP ViT$\dagger$} &  59.8 & 69.1 & 60.2 & 77.3 & 82.4 & 72.6 & \emph{No} \\ \midrule
    \rowcolor{gray!30} PFE (Ours) & 71.5 & 81.9 & 77.9 & 84.7 & 89.6 & 83.6 & \emph{Yes} \\
    \bottomrule[2pt]
    \end{tabular}}
    \end{minipage}
    \begin{minipage}[c]{0.5\linewidth}
    \label{tab:side_tuning}
    \centering
    \setlength{\tabcolsep}{1.8px}
    \small
     \resizebox{0.93\linewidth}{!}{
    \begin{tabular}{c|ccc|cc|cc}
    \toprule[2pt]
    \multirow{2}{*}{Settings} & \multicolumn{3}{c|}{DepthTrack} & \multicolumn{2}{c|}{LasHeR} & \multicolumn{2}{c}{VisEvent} \\ \cline{2-8}
    & F-score & Re & Pr & PR & SR & PR & SR  \\
    \midrule
    RGB Only & 58.5 & 59.9 & 57.2 & 58.5 & 45.6 & 72.7 & 55.9 \\
    Cross-Attn* & 59.9 & 59.7 & 60.0 & 61.5 & 49.3 & 74.4 & 57.5 \\
    Early Fusion* & 60.9 & 61.1 & 60.8 & 65.4 & 52.5 & 75.2 & 58.4 \\
    Extra Encoder & 60.2 & 60.2 & 60.2 & 63.7 & 51.8 & 75.7 & 58.8 \\
    \midrule
    w/o Zero Init & 62.6 & 62.6 & 62.6 & 70.1 & 55.9 & 77.1 & 60.2 \\
    Tuning UNet* & 62.3 & 62.4 & 62.2 & 69.3 & 55.5 & 76.7 & 59.8 \\
    \midrule
    Ours$\dagger$ & 65.5 & 65.2 & 65.8 & 71.0 & 56.9 & 78.5 & 61.9 \\
    \rowcolor{gray!30} Ours & 65.3 & 65.0 & 65.6 & 71.3 & 57.0 & 78.4 & 61.9 \\
    \bottomrule[2pt]
\end{tabular}}
    \end{minipage}
\end{table}

\noindent\textbf{Effectiveness of Multi-modal Sub-module Tuning.} We further validate the effectiveness of MST in Table~\ref{tab:pfe} \textbf{Right}. Model with modality-specific parameters is used.

\textit{Single Modal v.s. Multi Modal.} We directly evaluate the model trained after the first training stage with only RGB images for each benchmark. Without auxiliary modality, the F-score of DepthTrack decreases from 65.3\% to 58.5\%, validating that multi-modal sub-module tuning learns complementary features from the auxiliary modality.

\textit{Cross Attention, Early Fusion v.s. Sub-Module.} To validate the importance of the sub-module, we compare two alternative designs:
a) Injecting auxiliary modality information via cross-attention after extracting features with an extra image encoder. b) Fusing the noisy latent features of auxiliary and RGB images at the start of the UNet.
While these settings slightly improve performance over the single-modal baseline, they still underperform our design. Moreover, these approaches require tuning the UNet, which diminishes performance on RGB-N tracking.

\textit{Extra modality encoder v.s. VAE Encoder.} T2I-Adapter aligns the condition map with the RGB latent features using an extra modality encoder for conditional control image generation. We apply this method to inject auxiliary modality information, but it results in a decrease in PR on LasHeR from 71.3\% to 63.7\%.
Our method reuses the VAE encoder to generate noisy latent features for auxiliary modality images, which more effectively utilizes SD's knowledge.


\textit{Tuning UNet v.s. Freezing UNet.} After training on large-scale tracking datasets, the model adapts well to the tracking task. MST extends model's capability to integrate auxiliary modalities. Tuning UNet at the second stage leads to the forgetting of the learned knowledge, decreasing the F-score on DepthTrack from 65.3\% to 62.3\%.

\textit{W/ TNL2K v.s. W/O TNL2K.} In the first training stage, we add TNL2K into the training set to enable RGB-N tracking, leading to more training data than other multi-modal trackers. As shown in Table~\ref{tab:side_tuning}, the performance improvement is not related to these extra training data.

\begin{figure}
    \centering
     \includegraphics[width=1.0\linewidth]{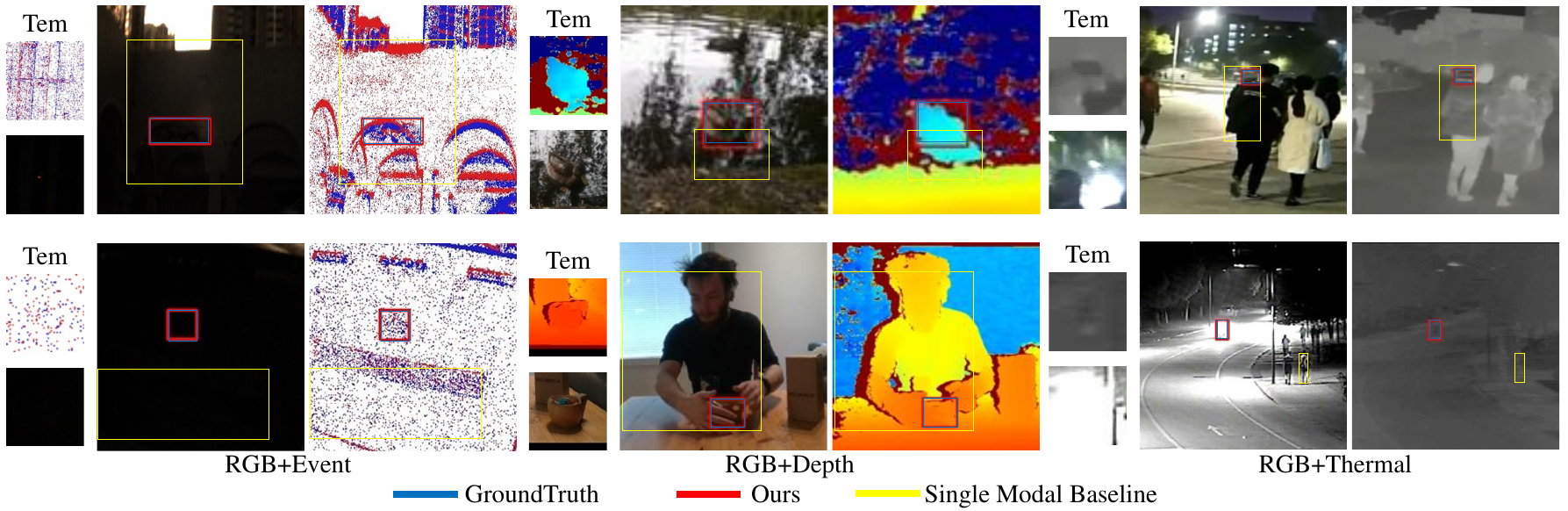}
  
     \caption{Visualization of tracking results.}
     \label{fig:modal_vis}
  \end{figure}

\noindent\textbf{Visualization.} To investigate if our method effectively learns complementary features from auxiliary modalities, we visualize some tracking examples in Fig.~\ref{fig:modal_vis}. The auxiliary modalities provide additional information under challenging conditions. For instance, object localization in RGB images becomes difficult under extreme lighting conditions. However, with event flow or thermal infrared input, the object remains visible in darkness or intense lighting. Our method, with auxiliary modality assistance, accurately distinguishes the object from the background, while the single-modal baseline fails.
\section{Conclusion}
We introduce Diff-MM, a tracker that unifies RGB-N/D/T/E tracking within a single parameter set. Unlike existing methods, which learn multi-modal knowledge only from limited multi-modal training data, our method explores another way to gain knowledge by leveraging the pre-trained Stable Diffusion model. To achieve this, a parallel feature extraction pipeline and multi-modal sub-module tuning method are proposed. Benefitting from extensive prior knowledge in SD, with only a single parameter set for all modalities, our method achieves the best performance across all multi-modal tracking benchmarks. The promising performance and extensive ablation studies demonstrate the effectiveness and reliability of our method.
{
    \small
    \bibliographystyle{ieeenat_fullname}
    \bibliography{main}
}

\end{document}